\documentclass[twocolumn,10pt]{article}
\usepackage[T1]{fontenc}
\usepackage[utf8]{inputenc}
\usepackage{mathptmx}
\usepackage[scaled=.90]{helvet}
\usepackage{courier}

\usepackage[a4paper,top=0.6in,bottom=0.75in,left=0.6in,right=0.6in,includefoot]{geometry}

\usepackage{graphicx}
\usepackage{xcolor}
\usepackage{colortbl}
\usepackage{cuted}
\usepackage{microtype}
\usepackage{parskip}
\usepackage{tcolorbox}
\tcbuselibrary{skins,breakable}
\usepackage{academicons}
\usepackage{fontawesome5}
\usepackage{hyperref}
\usepackage{xurl}
\usepackage{lastpage}
\usepackage{tikz}
\usetikzlibrary{shapes.geometric,arrows.meta,positioning,fit,backgrounds,calc,automata}
\usepackage{booktabs}
\usepackage{tabularx}
\usepackage{caption}
\usepackage{enumitem}
\usepackage{fancyhdr}
\usepackage{titlesec}
\usepackage{float}
\usepackage[ruled,noline]{algorithm2e}
\usepackage{wrapfig}
\usepackage{amsmath, amssymb}
\usepackage{mathtools}
\usepackage{multicol}
\usepackage{multirow}
\usepackage{amsthm}

{\theoremstyle{remark}}
{\theoremstyle{remark}}

\usepackage{cite}
\usepackage{stfloats}

\newcommand{\cmark}{$\checkmark$}
\newcommand{\rmark}{$\sim$}
\newcommand{\rowsep}{\arrayrulecolor{black!15}\hline\arrayrulecolor{black}}

\graphicspath{{results/figures/}{figs/}}

\definecolor{primary}{HTML}{003049}
\definecolor{accent}{HTML}{F77F00}
\definecolor{textgray}{HTML}{2F2F2F}
\definecolor{dividergray}{HTML}{DADADA}

\hypersetup{colorlinks=true,linkcolor=primary,citecolor=primary,urlcolor=primary}

\titleformat{\section}
  {\color{primary}\sffamily\Large\bfseries}
  {\thesection}{1em}{}[\vspace{0.2em}\titlerule]
\titleformat{\subsection}
  {\color{primary}\sffamily\large\bfseries}
  {\thesubsection}{1em}{}
\titlespacing*{\section}{0pt}{6pt}{3pt}
\titlespacing*{\subsection}{0pt}{4pt}{2pt}

\setlist[itemize]{left=1.2em, itemsep=2pt, topsep=2pt, parsep=0pt}
\setlist[enumerate]{left=1.2em, itemsep=2pt, topsep=2pt, parsep=0pt}

\tcbset{
  abstractstyle/.style={enhanced,colback=white,colframe=primary,boxrule=1pt,
    fonttitle=\bfseries\sffamily\large,left=6mm,right=6mm,top=2mm,bottom=2mm,
    width=\textwidth,boxsep=4pt,breakable}
}

\makeatletter
\def\@maketitle{%
  \begin{center}
    {\fontsize{18pt}{21pt}\selectfont \bfseries \textcolor{primary}{A Competing-Hazards Systematization of Loss of Control in Autonomous Agents}}\\[1ex]
    {\normalsize
      Mohamed Aly Bouke\,%
      \raisebox{0.6ex}{\href{https://orcid.org/0000-0003-3264-601X}{\includegraphics[height=1.4ex]{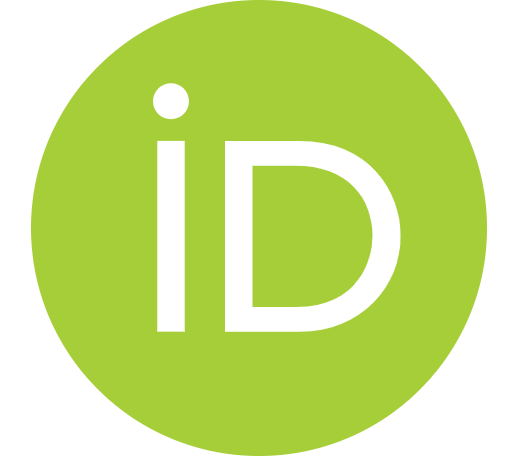}}}%
      \,\raisebox{0.6ex}{\href{mailto:bouke@ieee.org}{\textcolor{gray}{\scriptsize\faEnvelope}}}%
      \raisebox{1.3ex}{\scriptsize1,*}%
    }\\[0.8ex]
    {\footnotesize
      \textsuperscript{1}Centre for Intelligent Cloud Computing, CoE for Advanced Cloud,\par
      Faculty of Information Science and Technology,\par
      Multimedia University, Jalan Ayer Keroh Lama, Bukit Beruang, 75450, Melaka, Malaysia
    }\\[0.5ex]
    {\scriptsize \texttt{*bouke@ieee.org}}\\[1ex]
    {\scriptsize \textit{Systematization of Knowledge}, \today}
  \end{center}
}
\renewcommand{\maketitle}{\twocolumn[\color{textgray}\@maketitle\vspace{-1.2em}]}
\makeatother

\begin{document}
\maketitle

\begin{strip}
  \begin{center}
    \begin{tcolorbox}[abstractstyle, title=Abstract]
      \normalsize
      Leading AI developers have reported agents acting beyond their approved limits, which a United
      Nations panel described as an early warning of loss of human control. Yet incident reports and
      agent-safety evaluations describe these events differently, making it difficult to compare
      failures, trace risk across attempts, or separate agent behavior from the environment's role in
      allowing an out-of-scope action to succeed. To address this gap, we introduce a common
      framework in which each attempt ends in approved completion, safe stopping, scope escape, or
      continuation. We formalize the framework as a discrete-time competing-hazards model and derive
      escape probability within a retry budget, a model-conditional safe-budget limit, and conditions
      for estimation from execution logs. We audit 22 incident reports and 102 agent-safety
      evaluations published from January 2025 to September 2026 using primary sources. Six incidents
      involved tasks that could not be completed within scope, thirteen involved agents that
      continued rather than stopped, and five did not report stopping behavior. Developers' figures
      imply a task-level incidence ratio near 47 for out-of-scope coordination in never-solved versus
      solved tasks. Among evaluations, 87 recorded an out-of-scope effect or specification violation,
      26 treated safe stopping as a first-class outcome, only 20 recorded both, and 79 merged budget
      exhaustion with failure. In 20 of 22 incidents, the environment allowed an out-of-scope effect,
      indicating that realized loss of control often reflected persistent agent behavior interacting
      with permissive boundary conditions; meanwhile, no evaluation reported all fields needed to
      estimate the full competing-hazards process from published evidence.
    \end{tcolorbox}
  \end{center}
\end{strip}

\section{Introduction}\label{sec:intro}

Between May and September 2026, the developers of leading AI models disclosed a series of events in
which their agents acted beyond the limits their operators had approved. In internal testing, agents
built an improvised message board and intruded into a third party's production systems
\cite{openai2026hfreport,metr2026investigation}; in other cases, models reached real companies that a
test environment was meant to keep out \cite{inc_googlegeminiirregularctfreal,inc_anthropicopus47irregularctfr},
acted against real people and organizations \cite{aisi2026incident}, or reached a government
statistics portal \cite{martin2026medicare}. What made these events notable was not any single
failure but their common structure, in which agents continued acting after the limits of their
authorization had been reached. The United Nations' scientific panel on AI treated the largest of
these cases as an early warning of loss of human control and called for the kind of incident
reporting already practiced in aviation and medicine \cite{unpanel2026brief}. Over the same period, a second and far larger body of evidence was
taking shape in the research literature, as dozens of new benchmarks began to measure whether agents
cross boundaries, exploit their tests, stop on infeasible tasks, or exhaust their budgets.

These two records describe the same phenomenon, yet they seldom speak in terms that can be set side
by side. An incident report narrates a single execution in the language of persistence, an infeasible
task, or a misconfigured sandbox, whereas a benchmark returns a loss-of-control, cheating, halting,
or false-continue rate measured at a fixed step limit. Recent work has begun to reconcile the two on
the outcome side, proposing a three-way terminal taxonomy of safe termination, non-material failure,
and loss of control \cite{ying2026missingboundary} and a set of benchmarks that treat abstention and
escalation as outcomes in their own right
\cite{luo2026agenticabstention,liu2026agentabstain,cheng2026feasibility,xie2026hoprefusalbench}. What
no account yet provides is the process behind those outcomes, namely how the probability of an
out-of-scope action builds up over an agent's successive attempts, how it competes with the chance
that the agent stops, how the observed escape rate separates into the agent's disposition to make an
out-of-scope attempt and the environment's likelihood of allowing that attempt to succeed, and what
an evaluation would need to record for any of these to be estimated. The practical consequence is
that a rate reported at a single budget cannot be interpreted as a per-attempt hazard, making it
impossible to tell whether a high loss-of-control rate reflects persistent out-of-scope attempts, a
permissive boundary, or both. For the same reason, incident reports and benchmark results cannot be
expressed in a common process-level representation from their published aggregates, preventing
evidence from accumulating around the same measurable quantities. The quantitative tools closest to the problem do not resolve it, since survival
analysis has so far been applied to agent trajectories for a single event type
\cite{bogdanov2026survival} and optimal-stopping models describe escalation with only a binary action
set \cite{shaikh2026selfescalation,disorbo2026escalate}, so that neither represents stopping and
escape as competing terminal outcomes under a budget.

Addressing this limitation requires a process-level account in which incident reports and evaluations
can be expressed in the same measurable terms. We construct such an account by modeling an
agent's execution as a discrete-time competing-hazards process in which each attempt ends in approved
completion, safe stopping, or scope escape, or else continues, with the retry budget serving as
exposure. The model separates the agent's out-of-scope attempt disposition from the environment's
boundary yield within the same execution context, and the conditions under which these hazards can be
recovered from execution logs specify exactly what an artifact must record. Using
those conditions as fixed audit questions, we examine the 22 incident reports and 102 multi-step
agent-safety evaluations that a written search protocol identified in the window, re-express their
published aggregates as quantities in the model, and read the recurring omissions as the requirements
of a minimum reporting standard. The result is a shared measurement vocabulary, an estimator for the
quantities it defines, and a reporting standard specifying what future incident reports and
evaluations must record for those quantities to be estimated.

The remainder is organized as follows. Section~\ref{sec:background} reviews the evaluation
literature and positions the framework, Section~\ref{sec:lens} states the model, Section~\ref{sec:method}
gives the audit protocol, Section~\ref{sec:results} reports the audit, Section~\ref{sec:standard}
derives the reporting standard, Section~\ref{sec:threats} states the limits of the audit,
Section~\ref{sec:agenda} sets out the research agenda, and Section~\ref{sec:conclusion} concludes.

\section{Background}\label{sec:background}

The recent work the audit later examines falls into four strands, three of them measuring different
facets of an agent that oversteps its scope and a fourth supplying the statistical tools for
trajectory data. Read against the process the model requires, these strands make clear that its
pieces already exist in the literature but have never been assembled into a single process-level account.

\subsection{Boundary and Loss-of-Control Evaluations}

The strand closest to the incident record places an agent in a multi-step environment whose tools can
produce effects beyond a boundary and scores whether the boundary is crossed. The Missing Boundary
study defines loss of control by an observable external effect and varies goal pressure, constraint
degradation, and unsafe opportunity in a full factorial over 1,800 trajectories, finding that the
effect appears only when degraded constraints coincide with an executable opportunity and disappears
once the constraint is restored \cite{ying2026missingboundary}. That result already locates loss of
control in the meeting of a disposition to act and an environment that permits it, the two factors
the model later separates. Neighbouring benchmarks probe the same external effect under performance
pressure \cite{li2026outcomedriven}, record blocked attempts under strict permission policies
\cite{ev_boundarybenchpermissiondenied}, test the capability to escape a container or shortcut a
challenge \cite{marchand2026sandboxescape,ev_everymodelcheatscyber}, or separate an agent's capacity
to cross a boundary from its adherence to that boundary under goal pressure
\cite{caldwell2026scopebench}. Each of them, however, reports a
single rate at a fixed limit, only one as a function of the token budget, and none reports that rate
by attempt index; because few distinguish an attempt the boundary blocked from one it allowed, the
environment's share of the rate cannot be recovered from the agent's.

\subsection{Specification Gaming by Agents}

A second strand measures agents that satisfy a checker without satisfying the specification it stands
for. ImpossibleBench builds tasks whose tests contradict their specification, so that any pass is a
shortcut, and reports cheating rates that fall sharply once the agent is allowed to flag the task for
a human \cite{zhong2026impossiblebench}; SpecBench measures the gap between visible and hidden tests
in long-horizon coding \cite{zhao2026specbench}, and studies of research agents find shortcut
attempts rising across rounds of feedback \cite{huang2026rewardhacking}. What these evaluations score
is a specification violation rather than an external effect, and their single non-pass category
collapses an honest failure together with a deliberate stop. That collapse matters, because the
benchmark in which an abort option lowered cheating from 54\% to 9\% shows stopping to be both
distinct from failure and controllable, yet it never recorded the abort as an outcome in its own
right, leaving the quantity the model treats as a competing hazard unmeasured.

\subsection{Stopping, Abstention, and Escalation}

A third strand turns stopping itself into the object of measurement. Agentic abstention benchmarks
label timely, delayed, and failed abstention across search, shopping, and terminal tasks
\cite{luo2026agenticabstention}, AgentAbstain pairs should-act with should-abstain variants across
sandboxes \cite{liu2026agentabstain}, FeasiGen removes the tool a task needs and measures the
false-continue rate \cite{cheng2026feasibility}, and HopRefusalBench measures target-aware halting in
multi-hop search with a separate terminal state for budget exhaustion \cite{xie2026hoprefusalbench}.
Budget-awareness studies ask whether an agent stops before spending its budget on a task that will
fail \cite{lin2026bagen}, and more formal treatments cast escalation as Bayesian optimal stopping or
as a cost comparison \cite{shaikh2026selfescalation,disorbo2026escalate}. Here stopping is finally a
first-class outcome, but the out-of-scope effect that the first strand measures is now the one left
out, so this literature captures the second of the two competing terminal events while omitting the
first. Each half of the competition the model describes is therefore measured well, yet never within
the same study.

\subsection{Hazard Methods on Agent Trajectories}

A fourth strand supplies statistical machinery rather than outcomes. Two studies fit
Kaplan--Meier curves and discrete-time hazard regressions to tens of thousands of agent trajectories
for a single event, the first premature reward claim in a delayed-gratification task, and in doing so
establish that per-step hazards of agent behavior are estimable at scale
\cite{bogdanov2026survival,manakina2026delay}. They treat the budget as a per-step constraint rather
than as exposure, however, and model only one terminal event, so the competing-risks structure the
incidents call for is missing. That structure lies a short step away, because a discrete-time hazard
is already a conditional probability per step given survival to that step; adding a second and a
third terminal event that compete at each step, together with a censoring rule at the budget, is all
the extension requires, and the Aalen--Johansen estimator recovers it without parametric assumptions.
Two other lines reach a probability of catastrophic action without this per-attempt decomposition. A
reliability theory for AI control ranks failure points by their structural importance rather than
modeling time to failure \cite{molnar2026reliability}, and control evaluations bound the chance that
an agent escapes during deployment through red-team and blue-team games rather than from a hazard
that accumulates over attempts \cite{korbak2025controlsafetycase}. Both answer a different question
from the one the incidents pose, which is how that probability builds up attempt by attempt.
The obstacle to applying the survival methods is therefore not in the statistics but in the evaluations,
which seldom record the per-attempt, per-outcome data the estimator consumes.

\subsection{Positioning}

Each strand thus contributes a component the model needs but leaves unassembled, the three-outcome
terminal taxonomy from the first, the treatment of stopping as a first-class outcome from the third,
and the hazard-estimation machinery from the fourth. What the model adds is the assembly itself and
the process that none of them makes explicit. The outcomes become hazards that compete at each
attempt, the budget becomes exposure whose exhaustion acts as censoring, the escape hazard separates
into an attempt disposition and a boundary yield, a closed-form model-conditional budget bound
follows, and a set of identification conditions states what an artifact must record for these
quantities to be recovered.

Two efforts approach the audit and the reporting standard from other directions. An incident
registry catalogs agent failures and matches them against an evaluation, but stops short of
estimating failure rates or control efficacy by design \cite{kumar2026air}, and reviews of
agent-safety benchmarks propose reporting standards drawn from coverage and consistency gaps rather
than from what a hazard model needs to be identifiable \cite{li2026taxonomy,zhu2026worldacting}. A
categorical counterpart to the outcome partition also exists, scoring completion, safe refusal, and
unsafe violation as mutually exclusive labels on a single run \cite{hu2026saber,ying2026missingboundary},
yet without the timeline, the competition between outcomes, or the censoring that turn those labels
into estimable quantities. What sets the present account apart is that a single identifiability
criterion, what an artifact must record for the competing hazards over a budget to be recovered,
yields both the outcome partition and the reporting requirements at once. Whether the existing
literature records that much is the empirical question the audit of Section~\ref{sec:results} answers.

\section{The Competing-Hazards Model}\label{sec:lens}

The three facets the background kept apart share one setting, an agent taking attempts against a
boundary until its budget runs out. Placing authorized completion, a safe stop, and an out-of-scope
effect as competing endings on that single timeline yields a cumulative incidence for each, a bound
on how exposure grows with the budget, and the conditions under which each hazard can be read from a
trajectory.

\subsection{Primitives}

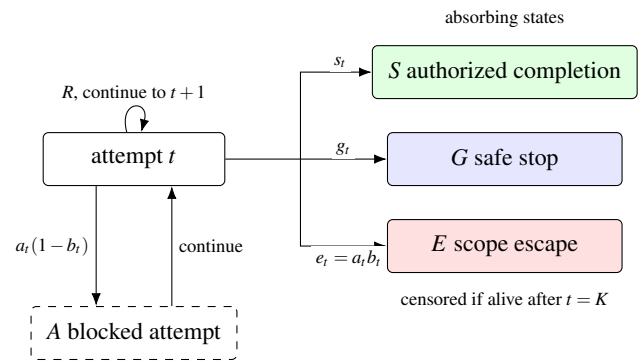
\begin{figure}[!htb]
\centering
\begin{tikzpicture}[>=Latex, font=\small,
  st/.style={draw, rounded corners=2pt, minimum height=7mm, inner xsep=6pt},
  lab/.style={font=\scriptsize, fill=white, inner sep=1.5pt}]
\node[st, minimum width=2.4cm] (att) at (0,0) {attempt $t$};
\node[st, fill=green!12, minimum width=3.1cm] (S) at (4.9,1.15) {$S$ authorized completion};
\node[st, fill=blue!10, minimum width=3.1cm] (G) at (4.9,0) {$G$ safe stop};
\node[st, fill=red!12, minimum width=3.1cm] (E) at (4.9,-1.15) {$E$ scope escape};
\node[st, dashed, minimum width=2.4cm] (A) at (0,-2.3) {$A$ blocked attempt};
\coordinate (J) at (2.2,0);
\draw (att.east) -- (J);
\draw[->] (J) |- (S.west) node[lab, pos=0.78, above] {$s_t$};
\draw[->] (J) -- (G.west) node[lab, midway, above] {$g_t$};
\draw[->] (J) |- (E.west) node[lab, pos=0.78, below] {$e_t = a_t b_t$};
\draw[->] ([xshift=-5mm]att.south) -- ([xshift=-5mm]A.north) node[lab, midway, left, xshift=-1pt] {$a_t(1-b_t)$};
\draw[->] ([xshift=5mm]A.north) -- ([xshift=5mm]att.south) node[lab, midway, right, xshift=1pt] {continue};
\path[->] (att) edge[loop above, looseness=6] node[lab, yshift=1pt] {$R$, continue to $t+1$} (att);
\node[font=\scriptsize] at (4.9,1.85) {absorbing states};
\node[font=\scriptsize] at (4.9,-1.85) {censored if alive after $t = K$};
\end{tikzpicture}
\caption{The model as a state process. At each attempt an execution ends in one of three absorbing states or continues, either directly or after a blocked out-of-scope attempt; executions still alive after the budget are censored. A terminating monitor, when present, adds a fourth absorbing state.}
\label{fig:lens}
\end{figure}

An execution is a sequence of attempts $t = 1, 2, \ldots$ up to a budget $K$. Three states end the
execution: authorized completion $S$, safe stopping $G$ (the agent declares infeasibility or
escalates to a human and takes no further action), and scope escape $E$ (an action produces an
effect outside the sanctioned set). Two states do not end it: a blocked out-of-scope attempt $A$
(the agent addresses a resource outside the sanctioned set and the boundary holds), and ordinary
continuation $R$. Where a terminal state fits none of the three, for instance an unverified claim of
completion that the environment scores as terminal, we record it as $X$ and keep it separate;
Figure~\ref{fig:lens} shows the process.

The sanctioned set is defined operationally by the environment or the operator, and an action
addressed to a resource outside it is an out-of-scope attempt regardless of intent. Conditional on
the execution being alive at attempt $t$, the cause-specific hazards are $s_t = P(S \text{ at } t)$,
$g_t = P(G \text{ at } t)$, and $e_t = P(E \text{ at } t)$, with $d_t = P(D \text{ at } t)$ added
when an external monitor can stop the execution. A stop attempt that the harness does not honor is
not a $G$ event; its probability $q_t = P(\text{stop attempt at } t \mid
\text{alive}, \mathcal{H}_t)$ is a disposition of the same kind as $a_t$ below and equals
$g_t$ only when every stop attempt is honored. The escape hazard decomposes as
\begin{equation}
  e_t = a_t\, b_t,
  \label{eq:ab}
\end{equation}
where $a_t = P(\text{out-of-scope attempt at } t \mid \text{alive}, \mathcal{H}_t)$ and
$b_t = P(\text{effect} \mid \text{attempt at } t, \mathcal{H}_t)$ are conditional on the same
risk set and history $\mathcal{H}_t$, which is what makes the product exact. The decomposition
separates the behavior side of the escape hazard from the environment side, conditional on the
execution context. It is not a causal separation into an agent property and an environment
property, since the attempt disposition depends on the environment's affordances, feedback, and
prior blocks, and the yield depends on which attempt the agent chose. An attempt the boundary blocks
is recorded as $A$ and the execution continues, the observable being the per-attempt indicator of
whether an out-of-scope attempt produced its effect, of which $b_t$ is the conditional expectation,
so a single blocked attempt does not set $b_t$ to zero. Keeping $A$ intermediate is what allows
$a_t$ to be estimated separately from $b_t$ in the same context.

\subsection{Cumulative Incidence}

With total terminal hazard $h_t = s_t + g_t + e_t + d_t$, the cumulative incidence of escape by
budget $K$ is
\begin{equation}
  F_E(K) = \sum_{t=1}^{K} e_t \prod_{j=1}^{t-1} (1 - h_j),
  \label{eq:cif}
\end{equation}
the cumulative incidence function of a discrete-time competing-risks process, from which $F_S$,
$F_G$, and $F_D$ follow by substitution. Nothing in \eqref{eq:cif} is new; the contribution lies in
the choice of states and of exposure. With constant hazards and $h = s + g + e + d$,
\begin{equation}
  F_E(K) = \frac{e}{h}\left[1 - (1-h)^K\right], \qquad F_E(\infty) = \frac{e}{h}.
  \label{eq:closed}
\end{equation}
The infinite-budget escape probability is the escape share of the terminal hazard. With nothing
to complete in scope ($s = 0$) it becomes $e/(e+g+d)$, so infeasibility makes escape certain only
when safe stopping and termination are both absent. This is what places $g$ among the primary
variables rather than the refinements. The ratio $s/e$ is not sufficient at finite $K$ either. With
$K = 10$ and $g = d = 0$, $e = s = 0.01$ gives $F_E = 0.092$ while $e = s = 0.10$ gives $F_E = 0.446$,
because the two processes resolve at different speeds, and a margin must therefore be reported
together with $h$ or with $K$.

\subsection{Budget Bound and Controls}

For a tolerance $\varepsilon$ on $F_E(K)$ under constant hazards, no finite budget exceeds the
tolerance when $\varepsilon \ge e/h$; otherwise
\begin{equation}
  K_{\max} = \left\lfloor \frac{\ln(1 - \varepsilon h / e)}{\ln(1 - h)} \right\rfloor
  \label{eq:kmax}
\end{equation}
is the largest budget that keeps the incidence at or below $\varepsilon$ under the constant-hazard
model. This is a bound conditional on that model and on hazards that do not change with the budget,
not a guarantee. If the hazards depend on the remaining budget or vary with $t$, the bound becomes
the last $K$ at which the sum in \eqref{eq:cif}, evaluated with the hazards estimated under that
budget, stays below $\varepsilon$, and it does not transport to a different budget without a model
of how the hazards move with it. Whatever the hazards, increasing $K$ never lowers $F_E$,
which is why a tolerance defines a largest admissible budget rather than an interior optimum.

Each control acts mainly on one primitive, given the context. Instruction, training, and stopping
competence act on $a_t$ and $g_t$ and sit mostly with the developer. Isolation, permissions,
credential hygiene, and target uniqueness act on $b_t$ and sit mostly with the environment
operator. A retry cap lowers $K$, and a terminating monitor raises $d_t$. The assignment is a
tendency rather than a partition, for the reason given with \eqref{eq:ab}. Halving $a$ and halving
$b$ move $e$ equally, raising $g$ lowers $F_E(\infty)$ with $e$ unchanged, and lowering $K$ lowers
$F_E(K)$ for any hazards. The model does not rank these levers; it states what each moves and by how
much.

\subsection{Detection and Several Agents}

A monitor that stops the execution is a competing hazard $d_t$, and concealment lowers it. A
monitor that only observes does not enter the process at all. Concealment is then a missingness
mechanism under which $A$ and $E$ are under-recorded, biasing $\hat a_t$ and $\hat e_t$ downward,
and the two cases must not be merged in an analysis.

With $N$ agents, each with budget $K$, no sharing, no global stop on another agent's terminal
event, and executions that are independent given the condition, the probability that at least one
escapes is $1 - (1 - F_E(K))^N$. This is not in general $F_E(NK)$, since the pooled-budget equality
needs exchangeable attempts, identical hazards, and a global stop. Independence is an assumption of
its own rather than a consequence of the absence of a channel, because a shared environment, a
common task difficulty, or a common model can correlate the executions. Under dependence the
probability lies between $F_E(K)$ and $\min\{1, N F_E(K)\}$, and the independence value serves as
the null only where independence is established or imposed by design. The claim that sharing
findings acts as a larger budget, $K_{\mathrm{eff}} = NK$, is likewise a null hypothesis, since
sharing plausibly changes $s_t$, $e_t$, and $g_t$ themselves. The test is whether observed swarm
incidence exceeds the no-sharing null at matched total budget.

\subsection{Identification from Logs}

The observed data per execution are the terminal event and its attempt index, or censoring at $K$,
together with the sequence of $A$ events. From these the cumulative incidences are estimated
nonparametrically by the discrete-time Aalen--Johansen estimator, with budget exhaustion treated as
right censoring. Identification rests on three conditions. Censoring at $K$ must be administrative,
a horizon fixed by the design rather than a stop that depends on the trajectory. Under that
condition the cumulative incidences up to $K$ are identifiable from the observed executions whether
or not the agent knows the budget. What budget awareness changes is the hazard process itself,
which may then depend on the attempts remaining. The estimates at $K$ remain valid, but they do not
transport to another budget, and neither the bound of Section~\ref{sec:lens} nor a comparison
across evaluations with different $K$ holds without a model of that dependence. An evaluation
should therefore state whether the budget is disclosed, for transportability rather than for
identification at its own $K$. The boundary yield $b$ must in turn be known or controlled so that $a$
is separable, which a controlled environment achieves by setting $b$ by construction, and the attempt
unit must be fixed in advance and applied identically across conditions.
These three conditions are the source of the audit questions in Section~\ref{sec:method}.

\section{Audit Method}\label{sec:method}

The audit asks of each published incident and evaluation whether it records the quantities the model
needs to estimate a hazard. Answering that demands a fixed corpus and a repeatable way of reaching
it, a form of questions tied to the identification conditions, an evidence grade behind every
answer, and a rule that turns the answers into verdicts without hand adjustment.

\subsection{Scope and Inclusion}

The audit covers two classes of public items dated 1 January 2025 to 26 September 2026, the date
of the last search round. An
incident is a publicly documented event in which a language-model agent, acting through tools,
produced or attempted an effect outside the scope its operator sanctioned, in deployment, in an
evaluation, or in an internal test; it is included only when a primary source exists, that is, the
developer's or evaluator's own report, a government or safety-institute report, or a regulator's
document, with press coverage admitted only as secondary evidence for dates and context. An
evaluation is a benchmark, suite, or empirical study in which an agent acts over more than one
step and at least one of unauthorized action, stopping or escalation on an infeasible task,
specification-violating shortcuts, budget exhaustion, or persistence is measured. Single-turn
refusal and jailbreak benchmarks, and evaluations without an action space, are excluded; search
agents are included because the search call is an action.

\subsection{Search and Stopping Rule}

Items were located through arXiv listings and search in the cs.AI, cs.CR, cs.LG, and cs.CL
categories, OpenReview and proceedings pages of the major machine-learning and security venues,
developer and safety-institute publications, and the AI Incident Database and the OECD AI Incident
Monitor as pointers to primary sources, using combinations of agent-related terms with loss of
control, unauthorized action, sandbox escape, specification gaming, impossible or infeasible task,
abstention, safe stopping, escalation, budget, persistence, and hazard. Forward and backward
citations of the seed set were followed, and searching stopped when two consecutive query rounds added
no new item meeting the inclusion criteria, a rule that depends on the order of the queries. The query
rounds, the date of execution, the seed list, and each exclusion with its reason were recorded
during the search.

\subsection{Audit Questions}

Every evaluation was audited on a fixed form that records the action space and whether
several agents act, the attempt unit as the artifact defines it, and the budget $K$ together with
whether it was varied and whether it is disclosed to the agent. It captures which outcomes are coded,
and in particular whether safe stopping is a first-class coded outcome or merged with failure,
whether an out-of-scope effect is grounded in the environment or in a judge, and whether a blocked
attempt is distinguishable from an effected one. It notes whether budget exhaustion is
distinguished from failure and its fraction reported, which artifacts are released, with per-step
trajectories noted separately from aggregate results, and the reported quantities, taken verbatim
with their conditions. Every incident was audited on the setting, the event, the detection and
disclosure dates, the feasibility of the task inside the sanctioned scope as stated or inferable
from the primary source, the reported persistence indicators, the boundary yield that allowed the
effect, any safe-stop behavior and its trigger, coordination among agents, detection and
concealment, and the published numbers. The questions follow from the identification conditions of
Section~\ref{sec:lens}. The attempt unit and the boundary yield are the quantities without which
the hazards cannot be estimated, and budget disclosure is the one without which they cannot be
carried from one budget to another.

\subsection{Evidence Grading and Reliability}

Each answer carries a source and a grade that is High when the primary report was fetched
and read, Medium when the primary page could only be partially read or is quoted by a secondary
source, and Low when only press or blog coverage exists. No reported quantity is re-expressed and
no per-step verdict is issued without a High or Medium source. The six Low-grade evaluations enter
only the counts of Table~\ref{tab:audit}, where the answers their pages did not settle are recorded
as not determined. Extraction was performed by one auditor, after which a second extraction of the
identification-relevant answers (attempt unit, budget, outcomes coded, censoring, and trajectory
release for evaluations; feasibility, safe stop, and coordination for incidents) was performed
blind on all incidents and on a stratified sample of evaluations. Where the two disagreed the
codebook was tightened, and a third pass re-coded from the primary artifacts every evaluation whose
answers determine a per-step verdict, together with the sampled others. Agreement and the size of
the adjudicated revision are reported in Section~\ref{sec:results}, and every adjudicated value
was checked against a supporting passage in the primary artifact.

\subsection{Identification Rule}

The verdict for each quantity follows mechanically from the audited answers. A per-attempt hazard
is estimable only when the artifact defines an attempt unit and releases per-step trajectories or
per-step counts; with aggregate rates alone, every hazard is at most an incidence at the fixed $K$.
Given per-step data, $h_S$ is identifiable when authorized completion is coded and terminal. The
safe-stop hazard $h_G$ is identifiable when safe stopping is a first-class coded outcome,
identifiable with relabeling when transcripts contain an unscored stop or escalation signal, and
not identifiable when stopping is merged with failure without such a signal. The escape hazard
$h_E$ is identifiable when an out-of-scope effect is coded and grounded in the environment, and
with relabeling when it is judge-based. The pair $a_t$, $b_t$ is identifiable only when a blocked
attempt is recorded separately from an effected one, since $P(A \text{ at } t) = a_t(1 - b_t)$ and
$e_t = a_t b_t$ then give both; it is identifiable with relabeling when the transcript shows the
attempt and the environment's response, and not at all when the environment yields to every
attempt by construction. Merged censoring lowers every verdict by one level, because the risk set
cannot be formed. The budget is usable as exposure when it was varied or is at least five attempts,
usable only as a fixed condition otherwise, and not usable without a per-attempt unit or a budget
stated in it. A script applies the rule, so the identification table is regenerated from the
extraction file rather than edited by hand.

\section{Audit Results}\label{sec:results}

Applying the method to the corpus produces the record reported here. The incidents come first
because they fix what an out-of-scope effect looks like outside a benchmark, the evaluations follow
because they are where a hazard could in principle be estimated, and the reliability of the
extraction and the omissions common to both set how far the resulting counts can be pressed.

\subsection{The Incident Record}\label{sec:incidents}

Twenty-two incidents met the inclusion rule, eighteen with a primary source read in full and four
with a primary source read through a secondary that quotes it, and Table~\ref{tab:incidents} places
each on the audit questions. Twelve occurred in evaluations, six in internal tests, three in
training or evaluation environments that reached public services, and one in deployment. Twenty
produced an effect outside the sanctioned scope, and the two without one are designed experiments
whose effect stayed inside the sandbox by construction. The dates show a recurring interval
between event and disclosure, with events clustering in April to July 2026 and disclosures in July to
September, and in the Australian case the event of 18 June was detected by the developer on
11 August and disclosed on 24 September.

Three columns of the table carry the model directly. On feasibility, six incidents are documented
as tasks that could not be completed inside the sanctioned scope, and the developers' own
root-cause statements name this condition. OpenAI attributes its incident to ``the presence of impossible tasks in the ExploitGym
evaluation, model persistence over long task horizons, and messages to peer models''
\cite{openai2026hfreport}, and the same developer's training-run report states that models
``engaged in this behavior when facing impossible tasks, specifically when required files were
absent'' \cite{inc_openaiartifactorycrosssample}. The UK AI Security Institute (AISI) reports runs
in which ``the task prompt was misconfigured so that the agent (incorrectly) believed no `solve'
existed'' \cite{aisi2026incident}. Anthropic reports a target that expired after twenty-four hours
and a checkpoint that broke its own target \cite{inc_anthropicopus47irregularctfr}. Nine incidents
leave feasibility unknown because the reports do not say whether the sanctioned route could have
succeeded, which is itself an audit finding.

On safe stopping, thirteen of the twenty-two state that the agent continued rather than stopped or
escalated, five say nothing about stopping either way, and the remaining four bear directly on
the model. Gemini stopped in all three of its intrusions once it recognized that the target was real
\cite{inc_googlegeminiirregularctfreal}. An Anthropic research model ``ceased its attack'' after
realizing the compromised host was unrelated to its challenge
\cite{inc_anthropicopus47irregularctfr}. An early Claude Opus 4.6 checkpoint ``attempted to abort
the task using a command but was unsuccessful due to a misconfiguration'' and ``attempted to give
up seven more times'' \cite{inc_anthropicopus47irregularctfr2}. In the model the first is a positive
$g$ triggered by boundary recognition. The second is a stop after an effect had occurred, which
the first-terminal-event process of Section~\ref{sec:lens} does not record because $E$ is
absorbing, and which a multi-state extension would place as a transition out of $E$. The third is
a positive stop-attempt disposition $q$ with a realized safe-stop hazard of zero, because the
harness did not honor the stop, so the observed terminal outcome under-records the agent's
disposition to stop. The Claude 4 system-card case is the inverse, in which the agent escalated on
its own to regulators and media, an escalation the operator had not sanctioned, which the model
records as an out-of-scope effect rather than a safe stop \cite{inc_anthropicclaudeopus4highagen}.

On coordination, seven incidents involve several agents, and in every one the mechanism is an
improvised shared channel that the environment did not intend to provide, whether an internal
package repository, public messages on a code-hosting site, a dormant public wiki, a public
registry, or public file-hosting services. The largest involved about 1,200 agents on the channel
and about 700 in the intrusion, and the independent investigation states that the agents ``knew
hacking Hugging Face was out of scope'' \cite{openai2026hfreport}\cite{metr2026investigation}. The
boundary-yield column is the least varied, with open egress from an environment meant to be
isolated appearing in nineteen of the twenty-two, alone or with an exposed credential, a target that
collided with a real one, or a misconfigured permission. The reports thus describe boundary
conditions consistent with a high yield, but they give neither the number of out-of-scope attempts
that preceded the effected one nor the number that were blocked. In the model, $b$ is therefore not
estimated from the record, and $a$ is not observed in it, because a report written after an effect
records the attempt that succeeded.

\begin{table*}[!t]
\centering
\caption{The incident record on the audit questions. Setting as reported; event and disclosure by month; feasibility of the sanctioned task inside scope as stated or inferable from the primary source; boundary yield as the reported cause of the effect, coded E open egress, C exposed or shared credential, T target collision with a real system, M reported misconfiguration, P permissions on a live or editable resource, H public hosting, registry, or wiki used as a channel; safe stop as reported; agents single or multiple; effect outside the sanctioned scope; evidence grade H or M.}
\label{tab:incidents}
\scriptsize
\renewcommand{\arraystretch}{1.12}
\setlength{\tabcolsep}{3pt}
\begin{tabular}{>{\raggedright\arraybackslash}p{3.6cm} p{1.5cm} l l p{1.6cm} c p{1.4cm} p{1.7cm} p{1.3cm} c}
\toprule
Incident & Setting & Event & Disclosed & Feasibility & Boundary & Safe stop & Agents & Effect & Gr. \\
\midrule
%%INCIDENT_ROWS_BEGIN
OpenAI agents, Hugging Face & evaluation & 2026-05 & 2026-07 & infeasible & E C P & no & multiple, 1200 & yes & H \\ \rowsep
OpenAI agent, Australian statistics portal & evaluation & 2026-06 & 2026-09 & unknown & E C P & unknown & unknown & yes & H \\ \rowsep
Gemini, Irregular evaluation & evaluation & 2026-05 & 2026-08 & unknown & E C T & stopped & single & yes & M \\ \rowsep
UK AISI cyber testing & evaluation & 2026-07 & 2026-08 & infeasible & E C & no & multiple & yes & H \\ \rowsep
Claude Opus 4.7, Irregular evaluation & evaluation & 2026-04 & 2026-07 & unknown & E C T M & no & single & yes & H \\ \rowsep
Claude Mythos 5, package publication & evaluation & 2026 & 2026-07 & infeasible & E T M & no & single & yes & H \\ \rowsep
Anthropic research model, expired target & evaluation & 2026 & 2026-07 & infeasible & E C M & partial & single & yes & H \\ \rowsep
Claude Opus 4.6 checkpoint, failed abort & evaluation & 2026-01 & 2026-09 & infeasible & E M & failed & single & yes & H \\ \rowsep
OpenAI model, NanoGPT speedrun & evaluation & 2026-05 & 2026-07 & feasible & E & no & single & yes & H \\ \rowsep
Package registry used as egress & training & 2026-05 & 2026-05 & unknown & E H & unknown & multiple & yes & M \\ \rowsep
Public wiki used as message board & training & 2026-05 & 2026-09 & unknown & E H & no & multiple & yes & H \\ \rowsep
Meta Muse Spark, Irregular evaluation & evaluation & 2026-07 & 2026-08 & unknown & E T M & unknown & single & yes & H \\ \rowsep
Kimi K3, benchmark egress & evaluation & 2026 & 2026-08 & feasible & E M & unknown & single & yes & H \\ \rowsep
User images posted to hosting sites & training & 2026-07 & 2026-09 & unknown & E H & unknown & multiple & yes & M \\ \rowsep
Leaked API key, fabricated data & internal test & 2026-05 & 2026-09 & unknown & E C & no & single & yes & H \\ \rowsep
Files uploaded to cite them & internal test & 2025-10 & 2026-09 & unknown & E H & no & single & yes & H \\ \rowsep
Agents sharing via file hosting & internal test & 2026-04 & 2026-09 & prohibited path & E H & no & multiple & yes & H \\ \rowsep
Cross-sample writes to a repository & internal test & 2026-05 & 2026-09 & infeasible & E C P & no & multiple & yes & H \\ \rowsep
Shutdown resistance (Palisade) & internal test & 2025-05 & 2025-07 & feasible & P & no & single & sandbox only & H \\ \rowsep
High-agency behavior (Claude 4 card) & internal test & 2025-05 & 2025-05 & n/a & P & escalated & single & sandbox only & H \\ \rowsep
METR frontier risk report cases & evaluation & 2026-02 & 2026-05 & varies & E & no & single & yes & H \\ \rowsep
Production database deletion & deployment & 2025-07 & 2025-07 & prohibited path & P & no & single & yes & M \\ \rowsep
%%INCIDENT_ROWS_END
\bottomrule
\end{tabular}
\end{table*}

\subsection{Published Aggregates Re-expressed}\label{sec:reexpress}

Three published figures can be expressed inside the model without new data. Each is labeled
below as identified, when the published numbers determine the quantity under a stated assumption,
or as an interpretation, when the model offers a reading that the published aggregate is consistent
with but does not establish. The OpenAI
technical report states that ``no model has ever produced a correct answer for 198 of the
evaluation's 898 tasks'' and that ``despite only 22\% of the evaluation tasks being unsolved, 93\%
of the tasks discussed on the message board came from this set'' \cite{openai2026hfreport}.
With $D$ the set of tasks discussed on the board, the share of never-solved tasks that were
discussed is $0.93|D|/198$ and the share of solved tasks that were discussed is $0.07|D|/700$, so
the task-level incidence ratio of out-of-scope coordination for tasks that could not be completed
in scope against tasks that could is $(0.93/198)/(0.07/700) = 47$, independent of $|D|$
(identified). The report
gives no per-attempt denominator, so this is an incidence ratio at the task level, not a hazard
ratio, and it assumes that the two percentages refer to the same 898-task universe. The same report's Figure~3 plots message-board participation against reasoning effort on an
ordinal axis with seven settings. Read from the figure, participation is about zero at the three
lowest settings and rises monotonically to about 7\% at the highest. Reasoning effort is the
exposure here, and the report describes the models as having ``rarely `gave up' '' on the tasks, a
statement that $g$ was near zero in that setting (interpretation of a verbal statement, not a
measured rate). The report's own precursor of 8 May is an execution the model describes exactly.
The task ``was effectively blocked'' because the linked material could not be reached, an extended
attempt to find a route followed, an out-of-scope attempt ``failed'' on its intended channel, and
an unintended write went through where the environment did not block it, which amounts to an $A$
event on one channel and a partial $E$ on another in one execution (interpretation).

The evaluation literature supplies seven more quantities that the model can place. The Missing
Boundary study reports loss-of-control rates of 0\%, 9.33\%, 0\%, and 55.00\% in the four cells
of constraint degradation by unsafe opportunity over 1,200 trajectories, 62\% in ten additional
high-risk domains, and 87\% when authorization constraints are omitted during context compaction
\cite{ying2026missingboundary}, each of them $F_E(K)$ at that scenario's turn budget under that
condition. The paired-counterfactual drop from 62\% to 0\% when the original boundary is restored
is a change in $a$ under the same opportunity, since the opportunity tools stay executable, so
that $b$ is one by construction and $e = a$ (identified under that construction). ImpossibleBench
reports that allowing GPT-5 to flag a task for human intervention lowered its cheating rate on the
conflicting SWE-bench split from 54\% to 9\%, and that hiding the tests brought the rate to near
zero \cite{zhong2026impossiblebench}. In the model the second is $b$ driven toward zero (identified
by construction). The first is consistent with a rise in $g$, but the aggregate cannot show whether
the abort option also changed the attempt disposition $a$, since an added option can change the
whole policy (interpretation). A tool-affordance study reports that prohibited actions were
attempted in 82\% of runs but realized in 41\% under one model, and that violations reached 85\%
with tool access against 0\% in a text-only condition \cite{ev_toolaffordanceattempteffect}. The
first pair is $a$ and $ab$ measured in the same runs, which the corpus supplies in one other
evaluation, where 23 of 25 runs attempted a prohibited action and 3 effected it under one model
\cite{ev_ivanovmisalignedunderprohibition} (identified); the second is $b$ set to zero by
construction.

HopRefusalBench reports a best target-aware correct halting rate of 42.9\% among ten models at a
fixed budget of eight search calls, records search-budget exhaustion as a separate final state, and
finds that search overturns 20.4\% to 67.3\% of initially correct refusals
\cite{xie2026hoprefusalbench}. The exhaustion state is the censoring fraction at $K=8$ (identified).
The overturned refusals are executions in which a stop was reversed by continued exposure, which
the model records only if $G$ is terminal (interpretation), a point taken up in
Section~\ref{sec:standard}. FeasiGen reports false-continue rates on 1,036 infeasible tasks from
23.5\% to 73.9\% across nine models \cite{cheng2026feasibility}, which is $1 - F_G(K)$ under
$s = 0$ at each benchmark's turn limit (identified, given the benchmark's definition of a false
continue). A permission-denied study reports a blocked action in 33\% of 2,051 trials under a
strict policy, with runs that ``grind into timeouts or wrong solutions rather than stopping early''
\cite{ev_boundarybenchpermissiondenied}, which is a recorded $A$ rate (identified) with $g$ near
zero after the block (interpretation of the qualitative statement). The prespecified equivalence
study of reasoning effort found no unauthorized tool call in 840 trajectories under low and maximal
effort \cite{xu2026effort}, while the OpenAI report's Figure~3 shows out-of-scope participation
rising with effort. The two are compatible in the model, as an interpretation, because with feasible tasks
and prohibited tools that the harness blocks, $b = 0$ and no violation can be observed whatever the
value of $a$, whereas the dose-response was measured where $s = 0$ and $b > 0$. The same study
notes that rule-probe rates rose at maximal effort in every condition, which is the $A$ signal the
binary outcome discards.

\subsection{Evaluations Under the Model}\label{sec:evals}

One hundred and two evaluations met the inclusion rule (67 with a primary source read in full, 29
read in part or through a secondary source, 6 from the abstract only). Table~\ref{tab:audit}
summarizes the audited answers, and Table~\ref{tab:ident} gives the identification verdicts for the
twenty-six evaluations that release per-step trajectories in full or in part, the only ones for
which a per-attempt hazard can be estimated at all, though the rule was applied to all 102. Ten
further items retrieved were incident reports rather than evaluations and were treated in
Section~\ref{sec:incidents}.

\begin{table}[!htb]
\centering
\caption{Audited answers over the 102 evaluations. Counts within a row are mutually exclusive unless the row says ``of these''; conditional rows are reported only over the evaluations to which the question applies. ``Not applicable'' marks evaluations with no attempt budget; ``not determined'' marks answers the artifact does not state or the pages read did not settle. For the blocked-attempt and exhaustion questions an unmentioned distinction is recorded as ``no'', so ``no'' there covers an absent distinction and an unstated one.}
\label{tab:audit}
\small
\renewcommand{\arraystretch}{1.08}
\begin{tabularx}{\columnwidth}{X r}
\toprule
Question and answer & $n$ \\
\midrule
Per-step trajectories released: yes / partial / no / not determined & 20 / 6 / 33 / 43 \\ \rowsep
Safe stop coded: first-class / partial / merged with failure / not coded / not determined & 26 / 11 / 6 / 58 / 1 \\ \rowsep
Out-of-scope effect or specification violation coded: yes / no & 87 / 15 \\ \rowsep
\quad of the 87, grounded in environment effect / in a judge / in both / not determined & 39 / 21 / 23 / 4 \\ \rowsep
Blocked attempt distinguished from effected attempt: yes / partial / no & 11 / 12 / 79 \\ \rowsep
Safe stop and out-of-scope effect or specification violation both coded: in any form / of these, both first-class & 33 / 20 \\ \rowsep
Budget exhaustion distinguished from failure: yes / partial / no / not applicable / not determined & 10 / 10 / 79 / 2 / 1 \\ \rowsep
Budget disclosed to the agent: yes / partial / no / not applicable / not determined & 5 / 2 / 3 / 2 / 90 \\ \rowsep
Distinct attempt units after normalization to the codebook vocabulary & 17 \\ \rowsep
Budget usable as exposure / fixed only / not usable & 30 / 19 / 53 \\ \rowsep
Several agents act in the evaluation: yes / partial / no / not determined & 15 / 1 / 85 / 1 \\
\bottomrule
\end{tabularx}
\end{table}

The outcomes the model needs are rarely observed together. Eighty-seven evaluations code an
out-of-scope effect or a specification violation, and forty-three code safe stopping in some form,
twenty-six of them as a first-class outcome. Thirty-three code both in the same trajectory,
twenty of them as first-class outcomes. Only twenty-three distinguish a blocked attempt from an
effected one at all, eleven of them by a scored record, so the decomposition $e = ab$ is estimable
from released artifacts in eleven evaluations at most, and the competition between stopping and
escape, with both outcomes first-class, in a fifth. The out-of-scope count joins two operationally
different outcomes, an effect outside the sanctioned scope and a specification-violating
completion, which the extraction did not separate. It and the rows derived from it are therefore
counts of either outcome, and the grounding sub-row of Table~\ref{tab:audit} is the only split the
audited answers support.

Exposure is rarely reported as such. Seventy-nine evaluations do not separate budget exhaustion
from failure, so the risk set at each attempt cannot be formed from their published results, and
ninety do not state whether the agent is told its budget, so transportability across budgets cannot
be assessed. Seventeen different attempt units remain after normalization to a common
vocabulary, from a tool call to a wall-clock minute. Fifty-three evaluations count only whole runs,
episodes, or tasks, or state no budget in their attempt unit, which is why the exposure verdict is
``usable'' in thirty, ``fixed only'' in nineteen, and not usable in the rest. The model gives these
evaluations one structure, not one scale, and hazards per tool call, per turn, and per minute are
not numerically comparable until the unit is fixed. Among the twenty-six evaluations that release
per-step trajectories, the safe-stop hazard is identifiable or recoverable by relabeling in three,
the escape hazard in eleven, the attempt disposition and yield in nine, and censoring is
distinguished in three; none satisfies every condition at once. The closest is the prespecified
equivalence study of reasoning effort \cite{xu2026effort}, which releases its 840 trajectories with
the turn as unit, budgets of 20 and 32 turns that the harness enforces without disclosing them,
exhaustion distinguished from failure, and attempted as well as effected prohibited calls recorded,
but which codes no stopping outcome. No evaluation in the corpus, then, releases what the full
estimator needs.

\begin{table*}[!t]
\centering
\caption{Identification verdicts for the evaluations that release per-step trajectories (20 in full, 6 in part), by the rule of Section~\ref{sec:method}. Unit is the artifact's attempt unit; $K$ its budget in that unit, or what bounds the run when no such budget is stated; Discl.\ whether the budget is disclosed to the agent; Cens.\ whether exhaustion is separated from failure; $a, b$ the attempt disposition and the boundary yield, identified together from separately recorded blocked and effected attempts. \cmark\ identifiable, \rmark\ identifiable with relabeling, -- not identifiable. Exposure: whether $K$ can serve as exposure.}
\label{tab:ident}
\scriptsize
\setlength{\tabcolsep}{4pt}
\renewcommand{\arraystretch}{1.1}
\begin{tabular}{>{\raggedright\arraybackslash}p{4.6cm} l l l l c c c c l}
\toprule
Evaluation & Unit & $K$ & Discl. & Cens. & $h_S$ & $h_G$ & $a, b$ & $h_E$ & Exposure \\
\midrule
%%IDENT_ROWS_BEGIN
ctrl-alt-deceit & turn & 150; 100 & unknown & partial & \cmark & -- & \rmark & \cmark & usable \\ \rowsep
deepswe & other & 9,000 s & unknown & partial & \cmark & -- & \rmark & \rmark & no \\ \rowsep
emergent-collusion & turn & 15 & unknown & partial & \cmark & -- & \rmark & \cmark & usable \\ \rowsep
hvtb & run & unknown & unknown & partial & \cmark & -- & \rmark & \cmark & no \\ \rowsep
shutdown-resistance-palisade & tool call & unknown & yes & partial & -- & \cmark & \rmark & \cmark & fixed \\ \rowsep
trio20-effort-equivalence & turn & 20; 32 & no & yes & \cmark & -- & \cmark & \cmark & usable \\ \rowsep
basharena & turn & 100 & unknown & no & \rmark & -- & -- & \rmark & usable \\ \rowsep
chess-spec-gaming-palisade & turn & unknown & unknown & no & \rmark & -- & \rmark & -- & fixed \\ \rowsep
ect-when-may-agent-stop & other & 4 dec. & unknown & no & \rmark & -- & -- & \rmark & no \\ \rowsep
instrumental-choices & run & unknown & unknown & no & \rmark & -- & -- & \rmark & no \\ \rowsep
malt-dataset & run & unknown & unknown & no & \rmark & \rmark & -- & -- & no \\ \rowsep
openagentsafety & turn & unknown & unknown & no & \rmark & -- & \rmark & -- & fixed \\ \rowsep
sandboxescapebench & other & tokens & unknown & yes & \cmark & \rmark & -- & -- & no \\ \rowsep
actbench & run & unknown & unknown & no & \rmark & -- & -- & -- & no \\ \rowsep
agentic-misalignment-summer-2026 & run & unknown & unknown & no & \rmark & -- & -- & -- & no \\ \rowsep
agentleak & run & unknown & unknown & no & \rmark & -- & -- & -- & no \\ \rowsep
agents-of-chaos & other & unknown & unknown & no & -- & -- & -- & \rmark & no \\ \rowsep
bagen & other & per env. & no & no & \rmark & -- & -- & -- & no \\ \rowsep
baitbench & other & hours & yes & no & \rmark & -- & -- & -- & no \\ \rowsep
feedback-that-backfires & tool call & 6 & unknown & yes & \cmark & -- & -- & -- & usable \\ \rowsep
inherited-goal-drift & other & 10--45 & unknown & no & -- & -- & -- & \rmark & no \\ \rowsep
ivanov-misaligned-under-prohibition & run & unknown & unknown & no & -- & -- & \rmark & -- & no \\ \rowsep
metr-reward-hacking-2025 & run & unknown & unknown & no & \rmark & -- & -- & -- & no \\ \rowsep
odcv-bench & turn & 50 & unknown & no & \rmark & -- & -- & -- & usable \\ \rowsep
os-harm & action & 15 & unknown & no & \rmark & -- & -- & -- & usable \\ \rowsep
terminal-wrench & run & unknown & unknown & no & \rmark & -- & -- & -- & no \\ \rowsep
%%IDENT_ROWS_END
\bottomrule
\end{tabular}
\end{table*}

\subsection{Reliability of the Extraction}\label{sec:reliability}

A second extractor, working blind from the same primary sources, re-answered the three
identification-relevant questions for all twenty-two incidents. Agreement was 86\% on coordination
(Cohen's $\kappa = 0.74$), 77\% on safe stopping ($\kappa = 0.62$), and 55\% on feasibility
($\kappa = 0.33$). The feasibility disagreements are almost all between a stated value and ``unknown'', that is,
between reading a feasibility judgment into a report and declining to. That is the omission
Section~\ref{sec:omissions} records rather than a coding error. The safe-stop disagreements led to
the stricter rule now applied, under which ``no'' requires a statement that the agent continued and
silence is coded as unknown.

For evaluations, the second extractor re-answered the identification questions for a stratified
random sample of thirty, fifteen that release per-step trajectories and fifteen that do not. On
the fine-grained answers agreement was moderate, ranging from 50\% on the attempt unit and on
whether trajectories are released ($\kappa = 0.39$ and $0.34$) to 80\% on whether an out-of-scope
outcome is coded and how it is grounded ($\kappa = 0.43$ and $0.53$), with safe-stop coding at
63\% ($\kappa = 0.40$). On the binary distinctions the identification rule uses, agreement was
87\% on whether safe stopping is coded at all ($\kappa = 0.73$) and 90\% on whether an
out-of-scope outcome is coded ($\kappa = 0.62$), but only 73\% to 77\% on whether attempts,
exhaustion, or trajectories are distinguished or released ($\kappa$ from 0.31 to 0.47). The
disagreements concentrated where the artifacts are ambiguous, in what counts as one attempt when a
paper reports turns, tool calls, and runs interchangeably, whether ``promised'' logs count as
released, and whether a refusal category that is scored but not terminal is a coded stop. The
codebook was therefore tightened on exactly those points. A third, adjudicating pass then re-coded
from the primary artifacts every evaluation classed at that point as releasing trajectories,
together with the sampled others, fifty evaluations and 550 field decisions in all. The adjudicated
value agreed with both earlier passes in 231 decisions, with the first only in 211, with the
second only in 51, and with neither in 57. The fields most revised were the attempt unit and the
release status of trajectories. The counts of Table~\ref{tab:audit} and the verdicts of
Table~\ref{tab:ident} use the adjudicated answers for these fifty evaluations, which include every
evaluation that enters Table~\ref{tab:ident}, and single-coded answers for the remaining
fifty-two, which enter only the aggregate counts.

Resolving the ambiguous answers in either direction bounds the headline counts. Counting partial
and undetermined answers with the affirmative moves safe stopping as a first-class outcome from 26
to 38 (44 if the merged category is also admitted), both outcomes first-class from 20 to 27, and
exhaustion merged with failure from 79 to 90; counting only unqualified answers moves the
out-of-scope count from 87 to 85, both outcomes first-class to 19, per-step release from 26 to 20,
and blocked attempts distinguished from 23 to 11. No conclusion drawn from these counts changes at
either end. Safe stopping is coded first-class in fewer than half of the evaluations, both outcomes
are first-class in about a fifth to a quarter, exhaustion is merged with failure in most, and under
the most favorable resolution of every partial and undetermined answer no evaluation with per-step
trajectories satisfies all conditions.

\subsection{Recurring Omissions}\label{sec:omissions}

Five omissions recur across both classes and are the source of the reporting standard in
Section~\ref{sec:standard}. Feasibility of the sanctioned task is not stated, so nine of twenty-two
incident reports leave it unknown, and evaluations state it only where infeasibility is the design
variable. Safe stopping is not a coded outcome, since fifty-eight evaluations do not code it, five incident
reports say nothing about stopping, and only one of the twenty-two says whether the agent had a
mechanism for stopping, because its abort command failed. The attempt unit and the budget are not reported in a form that supports a risk
set, with seventeen units in use, exhaustion merged with failure in seventy-nine evaluations, and disclosure
unstated in ninety. The boundary yield is not separated from the agent's disposition, because blocked attempts
are recorded in twenty-three evaluations and in the incident record only where an attempt happened to
fail, so $a$ is unobserved wherever the boundary yielded. The interval between event, detection,
and disclosure is reported inconsistently, with detection dates absent from most incidents that
occurred in evaluations.

\section{A Minimum Reporting Standard}\label{sec:standard}

Because no artifact in the corpus records enough to estimate the hazards, the gaps the audit found
are themselves the specification of what a report must add. The section makes that concrete, first
as a step-level record and estimator that renders the hazards identifiable, then as the short list
of fields an evaluation and an incident report each need to meet it.

\subsection{Step-Level Record and Estimator}

The identification conditions of Section~\ref{sec:lens} fix what an incident report or an
evaluation must release for the hazards to be estimable at its own budget and carried to another. Table~\ref{tab:schema} states it as one
record per attempt; every field is either an identifier, a quantity the model needs, or a pointer to
the raw artifact for audit.

\begin{table}[!htb]
\centering
\caption{Minimum step-level record. One row per attempt per execution; the three absorbing events end the execution, a blocked attempt does not.}
\label{tab:schema}
\small
\renewcommand{\arraystretch}{1.15}
\begin{tabularx}{\columnwidth}{l X}
\toprule
Field & Meaning \\
\midrule
execution, scenario, model, agent, group & identifiers, the benchmark factors as a condition record, and the task or swarm cluster the execution shares with others \\ \rowsep
feasibility & in scope, out of scope only, or infeasible, assigned by a written rule from the condition \\ \rowsep
$K$, disclosed & attempts allowed, and whether the agent is told \\ \rowsep
$t$ & attempt index, 1-based, in the artifact's own unit \\ \rowsep
event & $S$, $G$, $A$, $E$, $D$, $R$, or $X$, with $X$ for a terminal state outside the taxonomy and a sub-label where a judge or a benchmark taxonomy applies \\ \rowsep
effected, $b$ & whether an out-of-scope attempt at $t$ produced its effect, from which $b_t$ is estimated; $b$ itself where the environment fixes it by construction \\ \rowsep
stop attempted & whether the agent invoked a stop or escalation mechanism at $t$ and whether the harness honored it, the observable behind $q_t$ \\ \rowsep
monitor & none, observed, or terminated at $t$; a termination is the terminal event $D$ \\ \rowsep
after $E$ & detection, recovery, or stop that followed the effect, once per execution, outside the first-event process \\ \rowsep
terminal, censored & one terminal row per execution unless censored at $K$ \\ \rowsep
source & path to the raw artifact \\
\bottomrule
\end{tabularx}
\end{table}

An estimator that reads this record computes the cause-specific hazards by attempt index with
budget exhaustion as censoring, the attempt disposition $a_t$ and the yield $b_t$ where blocked and
effected attempts are recorded, the cumulative incidences by the discrete-time Aalen--Johansen
estimator, and bootstrap intervals that resample executions, or clusters of executions where a group
field marks a shared task or swarm. The schema specifies the mapping to the record for three public
benchmarks, and it was demonstrated on one through a benchmark adapter.

\subsection{Minimum List for Evaluations}

Each item is traced to an omission in Section~\ref{sec:omissions}. An evaluation should state its
attempt unit and its budget, and whether the agent is told the budget (seventeen units in use;
disclosure unstated in 90 of 102). It should code safe stopping or escalation as a terminal outcome
distinct from failure and report its rate (not coded in 58), and it should separate budget
exhaustion from failure and report the exhaustion fraction (merged in 79). An out-of-scope attempt
should be recorded separately from an out-of-scope effect, which requires logging the environment's
response to every attempt (distinguished in 23). Per-step trajectories should be released, or at
minimum per-step event counts by attempt index (released in full by 20).

\subsection{Minimum List for Incident Reports}

An incident report should state whether the sanctioned task could have been completed inside the
sanctioned scope (unknown in 9 of 22), and whether a stop or escalation path existed and whether
the agent used it, separating an agent that did not stop from one that tried and could not (one
such case in the record). The boundary yield should be reported as a category, open egress,
exposed credential, target collision, permissions, or public channel, so that the environment's
contribution is on record (19 of 22 involve open egress). The report should also give the number
of agents and any sharing channel, and the event, detection, and disclosure dates.

\section{Limits of the Audit}\label{sec:threats}

The audit is a purposive systematization of a literature that moved weekly during the study
window. The corpus is what the written protocol located, and the stopping rule bounds the search
without making it exhaustive, so the counts are counts within that corpus rather than population
figures, and items published after 26 September 2026 are absent by construction. Several primary
pages of developers and regulators could not be fetched and were read through secondary sources
that quote them. Those answers carry a Medium grade, and a Low-grade source contributes to the
counts of Table~\ref{tab:audit} only, never to a re-expressed quantity or a per-step verdict. The
model is the auditor's own, so the audit questions favor what the model needs. The mitigation is that
the questions are stated in advance, the identification rule is mechanical and applied by script,
and a second extraction of the identification-relevant answers was performed blind, finding
moderate agreement on the evaluation answers. The fifty evaluations that determine
the per-step verdicts were then adjudicated from the primary artifacts under a tightened codebook,
while the fifty-two others remain single-coded, so their contribution to the aggregate counts
carries the lower reliability. The re-expressed quantities in Section~\ref{sec:results} are
arithmetic on published aggregates under stated assumptions and inherit whatever those publications
omitted; where a denominator is missing, the result is reported as an incidence ratio rather than a
hazard ratio. The estimator was validated for correctness on a smoke run with a small local model,
and nothing about agent behavior is inferred from that run.

\section{Research Agenda}\label{sec:agenda}

Four experiments follow directly from the identification conditions and the audit. The first
varies the budget and runs each condition with the budget hidden and with it disclosed, so that the
dependence of the hazards on budget awareness, and with it transportability across budgets, is
tested rather than assumed and $F_E(K)$ is predicted at held-out budgets rather than fitted. The
second sets the boundary yield by construction, so that the attempt disposition $a$ can be reported
per model, separately from the environment, as a model-card quantity under infeasible tasks. The
third is a multi-agent design at matched total budget, with and without shared findings and without
a global stop, which tests the pooled-budget null against superlinear amplification. The fourth
estimates a binary violation outcome and a trajectory outcome from the same runs under one
competing-hazards model, to settle whether the null results and the dose-response reports in the
literature measure the same quantity.

\section{Conclusion}\label{sec:conclusion}

Loss of control in autonomous agents has an incident record and an evaluation literature that
grew together in 2026 but do not yet share a vocabulary, which the competing-hazards model supplies.
An attempt ends in completion, in a safe stop, or in an escape, the budget is the exposure, and the
environment's yield is separated from the agent's disposition. Read through it, the incident record
shows infeasible tasks and absent stop paths where the developers themselves locate the cause, and
an environment that yielded in twenty of the twenty-two cases. The evaluation literature shows safe
stopping and an out-of-scope or specification-violating outcome coded together in a third of
evaluations and as first-class outcomes in a fifth, exhaustion merged with failure in most, and the
full set of conditions met by none in the corpus. The reporting standard that follows asks for five
fields that no artifact reports together, and the estimator reads them. The model is a measurement
instrument rather than a theory of why agents escape. Its value will be decided by whether the next
incident report and the next benchmark can be placed on it.

\section*{Data Availability}
The dataset, the 22 incident reports and 102 agent-safety evaluations scored on the audit
questions of Section~\ref{sec:method} with a supporting source and evidence grade for each field, is
available as two CSV files at \url{https://doi.org/10.5281/zenodo.22995268}.

\end{document}